\documentclass{article}

\usepackage[preprint]{corl_2022} %

\title{SAFER: Data-Efficient and Safe Reinforcement Learning via Skill Acquisition}

\usepackage{algorithmic}
\usepackage{microtype}
\usepackage{selectp}
\usepackage{enumitem}
\usepackage{amsfonts}
\usepackage{amssymb}
\usepackage{amstext}
\usepackage{amsmath}
\usepackage{xspace}
\usepackage{theorem}
\usepackage{graphicx}
\usepackage{algorithm}
\usepackage{booktabs}
\usepackage{url}
\usepackage{graphics}
\usepackage{colordvi}
\usepackage{import}
\usepackage{caption}
\usepackage{multirow}
\usepackage{subcaption}
\usepackage{wrapfig}
\usepackage{mathtools}
\usepackage{pgfplots}
\usepackage{wrapfig}
\usepackage{sidecap}

\pgfplotsset{compat=1.16}

\usepackage{amsmath,amsfonts,bm}

\def\eqref#1{equation~\ref{#1}}
\def\Eqref#1{Equation~\ref{#1}}

\def\floor#1{\lfloor #1 \rfloor}
\def\1{\bm{1}}

\def\va{{\bm{a}}}

\def\vc{{\bm{c}}}

\def\vs{{\bm{s}}}

\def\vz{{\bm{z}}}

\DeclareMathAlphabet{\mathsfit}{\encodingdefault}{\sfdefault}{m}{sl}
\SetMathAlphabet{\mathsfit}{bold}{\encodingdefault}{\sfdefault}{bx}{n}

\newcommand{\qed}{\hfill\blacksquare}

\newtheorem{theorem}{Theorem}[section]

\newtheorem{proposition}[theorem]{Proposition}
\newtheorem{assumption}{Assumption}[section]

\newif\ifcomments

\ifcomments
    \providecommand{\yinlam}[2][]{{\protect\color{violet}{[\textbf{yinlam}:\textbf{#1} #2]}}}
    \providecommand{\dylan}[2][]{{\protect\color{blue}{[\textbf{dylan}:\textbf{#1} #2]}}}
    \providecommand{\nevan}[2][]{{\protect\color{brown}{[\textbf{nevan}:\textbf{#1} #2]}}}
    \providecommand{\bo}[2][]{{\protect\color{red}{[\textbf{bo}:\textbf{#1} #2]}}}
\else 
    \providecommand{\yinlam}[2][]{}
     \providecommand{\dylan}[2][]{}
     \providecommand{\nevan}[2][]{}
     \providecommand{\bo}[2][]{}
\fi

\pgfmathdeclarefunction{gauss}{2}{%
  \pgfmathparse{1/(#2*sqrt(2*pi))*exp(-((x-#1)^2)/(2*#2^2))}%
}

\author{
  Dylan Slack\thanks{Work performed while an intern at Google AI}\\
  UC Irvine\\
  \texttt{dslack@uci.edu}
  \And
  Yinlam Chow \\
  Google Research \\
  \texttt{yinlamchow@google.com}
  \And
  Bo Dai \\
  Google Research \\
  \texttt{bodai@google.com}
  \And
  Nevan Wichers \\
  Google Research \\
  \texttt{wichersn@google.com}
}

\newcommand\blfootnote[1]{%
  \begingroup
  \renewcommand\thefootnote{}\footnote{#1}%
  \addtocounter{footnote}{-1}%
  \endgroup
}

\begin{document}
\maketitle

\begin{abstract}
Methods that extract policy primitives from offline demonstrations using deep generative models have shown promise at accelerating reinforcement learning (RL) for new tasks.
Intuitively, these methods should also help to train \textit{safe} RL agents because they enforce useful skills.
However, we identify these techniques are not well equipped for safe policy learning because they ignore negative experiences (e.g., unsafe or unsuccessful), focusing only on positive experiences, which harms their ability to generalize to new tasks safely.
Rather, we model the latent \textit{safety context} using principled contrastive training on an offline dataset of demonstrations from many tasks, including both negative and positive experiences.
Using this latent variable, our RL framework, SAFEty skill pRiors~(SAFER) extracts task specific safe primitive skills to safely and successfully generalize to new tasks.
In the inference stage, policies trained with SAFER learn to compose safe skills into successful policies. We theoretically characterize why SAFER can enforce safe policy learning and demonstrate its effectiveness on several complex safety-critical robotic grasping tasks inspired by the game Operation,\footnote{\href{https://en.wikipedia.org/wiki/Operation_(game)}{\texttt{https://en.wikipedia.org/wiki/OperationGame}}} in which SAFER outperforms state-of-the-art primitive learning methods in success and safety.
\end{abstract}


\section{Introduction}
\label{sec:introduction}

\blfootnote{\textit{Decision Awareness in Reinforcement Learning Workshop at the 39th International Conference on Machine Learning (ICML)}, Baltimore, Maryland, USA, 2022. Copyright 2022 by the author(s).}Reinforcement learning (RL) has demonstrated strong performance at solving complex control tasks.
However, RL algorithms still require considerable exploration to acquire successful policies.
For many complex safety-critical applications (i.e., autonomous driving, healthcare), extensive interaction with an environment is impossible due to dangers associated with exploration.
These difficulties are further complicated by the challenging nature of specifying safety constraints in complex environments.
Nevertheless, relatively few existing safe reinforcement learning algorithms can \emph{rapidly} and \emph{safely} solve complex RL problems with hard to specify safety constraints.

One promising route is offline primitive learning methods~\citep{Singh2021ParrotDB, pertsch2020accelerating,pertsch2021guided,Ajay2021OPALOP}.
These methods use offline datasets to learn representations of useful actions or \textit{behaviors} through deep generative models, such as normalizing flow models or variational autoencoders (VAE).
Specifically, they treat the latent space of the generative model as the abstract action space of higher-level actions (i.e., skills). These methods train an RL agent to map states onto the abstract action space of skills for each downstream task using the learned primitives. This approach can significantly accelerate policy learning because the generative model learns useful primitives from a dataset, simplifying the action space \citep{dulac2015deep}.

\begin{wrapfigure}{r}{0.45\textwidth}
    \centering
        \scalebox{0.65}{
        \input{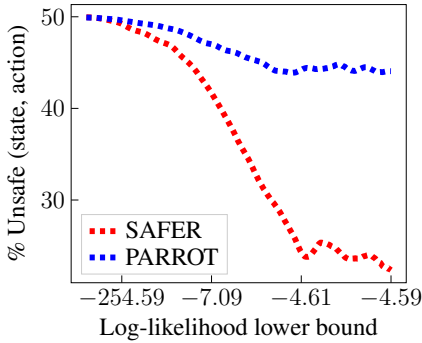}}
    
  \caption{
  \textbf{Evaluating the concentration of unsafe data in high likelihood regions} by computing the \% of unsafe state-action pairs in a holdout dataset of a safe robotic grasping task.
  PARROT assigns high likelihoods to unsafe data, i.e., it does not encourage safety, while SAFER has much lower likelihood in unsafe data, so it will encourage safety.
 }
  \label{fig:tradoffexample}
\end{wrapfigure}

However, primitive learning techniques suffer from a critical drawback when applied to safety concerned tasks.
Intuitively, if trained on datasets consisting of trajectories that are both safe and successful, offline skill learning methods should capture \emph{safe and useful} behaviors and encourage the \emph{rapid} acquisition of safe policies on future tasks (\textit{downstream} learning).
For example, when trained on data from everyday household tasks, these methods should learn behaviors that successfully and safely accomplish similar tasks, such as handling objects carefully or avoiding animals in the environment.
However, when offline skill learning methods are trained only with safe experiences, the unsafe data is out of the training distribution.
It is well known that deep generative models have problems generalizing to out of distribution data, which increases the likelihood of unsafe actions (see Fig.~\ref{fig:tradoffexample}) \citep{knowwhattheydontknow, classcond, whyfailood}.
\textit{Thus, current state of the art primitive learning techniques may, counter-intuitively, encourage unsafe behavior}.

In this work, we identify that modeling the latent \textit{safety context} is the key to overcoming these challenges.
To this end, we introduce \textit{SAFER}: \underline{safe}ty skill p\underline{r}iors, a primitive learning technique that \emph{accelerates} reinforcement learning with \emph{safe} actions. (An overview is provided in Figure~\ref{fig:overview}.)
To acquire safe skills, SAFER \textbf{i)} uses a contrastive loss to distinguish safe and unsafe data and \textbf{ii)} learns a posterior sampling distribution of a latent safety variable, that captures different
safety contexts. 
Using the safety context, SAFER established a set of task specific safe actions, greatly improving safety generalization.
Consequently, policies trained using the SAFER abstract actions as the action space will learn to compose a set of safe policy primitives.
As shown in Figure~\ref{fig:tradoffexample}, SAFER assigns much lower likelihood to unsafe states and actions, indicating that it will better promote safe behaviors when applied to downstream RL.
To demonstrate the effectiveness of SAFER, we evaluate it on a set of complex safety-critical robotic grasping tasks.  When compared with state of the art primitive learning methods, SAFER has both a higher success rate and fewer safety violations.

\vspace{-4mm}
\section{Related Work}
\label{sec:related_work}
\vspace{-2mm}

\textbf{Safe Exploration} Several related works focus on safe exploration in RL when there is access to known constraint function
\citep{wachi2019constrainedmarkov, constrained_policy_optimization, continuous_action_spaces, Bharadhwaj2021ConservativeSC,  Narasimhan2020ProjectionBasedCP, Yang2020AcceleratingSR, chow2018lyapunov, chow2019lyapunov, Achiam2019BenchmarkingSE, NIPS2017_766ebcd5, 7526658, NEURIPS2020_8df6a659}.  
In our work, we focus on the setting where the constraint function cannot be easily specified and must be inferred \textit{entirely} from data, which is critical for scaling safe RL methods to the real world.
To this end, a few works consider a similiar setting where the constraints must be inferred from data. \citet{Thananjeyan2021RecoveryRS} uses an offline dataset of safety constraint violations to learn about safety constraints and trains a policy to recover from safety violations so the agent can continue exploring safely. \citet{natural_language_constraints} use natural language to enforce a set of safety constraints during policy learning.
Compared to our work, these methods focus on constrained exploration in a single task setting.
Instead, we consider accelerating learning across multiple tasks through learned safe primitives.

\textbf{Demonstrations for Safe RL} 
The use of demonstrations to ensure safety in RL has received considerable interest in the literature \citep{Rosolia2018LearningMP, Thananjeyan2021ABCLMPCSS, Driessens2004IntegratingGI, practicalrl}.
Most relevant, \citet{Srinivasan2020LearningTB} use unsafe demonstrations to constrain exploration to only a safe set of actions for task adaption. 
\citet{Thananjeyan2020SafetyAV} relies on a set of sub-optimal demonstrations to safely learn new tasks.
Though these works leverage demonstrations to improve safety, they each rely on task specific demonstrations. 
Instead, we focus on learning generalizable safe primitives, which we transfer to downstream tasks, and we demonstrate this can greatly accelerate safe policy learning.

\textbf{Skill Discovery} Various works consider learning skills in an online fashion \citep{Eysenbach2019DiversityIA, nachumnearopptimal2019, Sharma2020DynamicsAwareUD, latentskillplanning, NIPS2009_e0cf1f47}. 
These methods learn skills for planning \citep{Sharma2020DynamicsAwareUD} or online RL \citep{Eysenbach2019DiversityIA, nachumnearopptimal2019}.
In contrast, we focus on a setting with access to an offline dataset, from which the primitives are learned. 
Further works also use offline datasets to extract skills, and transfer these to downstream learning \citep{pertsch2020accelerating, pertsch2021guided, Ajay2021OPALOP}, but they do not model the safety of the downstream tasks, which we demonstrate is critical for safe generalization.

\textbf{Hierarchical RL} Numerous works have found learning high level primitives using auxiliary models and controlling these with RL beneficial \citep{Singh2021ParrotDB, xuebinpengnips19, chandak19aleanringaction, NEURIPS2018_e6384711, hausum2017multimodal, florensa2017stochastic, Fox2017MultiLevelDO, maxqdietterich, context_vars}.
Though these works propose methods that are capable of accelerating the acquisition of successful policies, they do not specifically consider learning with safety constraints, which makes them susceptible to the generalization issues discussed in Section~\ref{sec:introduction} and Section~\ref{sec:background}, where these methods can inadvertently make unsafe behavior high likelihood. In contrast, SAFER learns a hierarchical policy that explicitly considers the safety of tasks, resulting in both safe and successful generalization to downstream tasks, addressing the aforementioned issues.

\vspace{-3mm}
\section{Background}
\label{sec:background}
\vspace{-1mm}

In this section, we provide background for our problem setting. 
Recall the motivating household robotics example where we wish to train an agent to accomplish a series of household tasks.
The agent must learn to do tasks like set a cast iron pot to boil, remove dirt off a dish with a sponge, or cut an apple with a knife.
Within these tasks, there are different goals and notions of safety.
For example, the robot can safely drop the sponge but cannot safely drop the cast iron pot while cooking, because this would be quite dangerous.
From a training perspective, it is difficult to devise safety violation functions for all tasks, given how many ways one could behave unsafely with a cast iron pot or knife.
However, it is straightforward to determine whether the task is successful (e.g., the apple is cut in half or it isn't).
Consequently, it is more reasonable to assume an \textit{offline data collection process} where a large set of behaviors have been annotated for success and safety violation, through simulation or real world experience, and agents must rely entirely on the existing data to learn safety constraints when generalizing to downstream tasks, though they may have access to a sparse reward signal.

\textbf{Safety MDP} In a setting with different tasks and safety constraints, for each task $\mathcal{T}$, the agent's interaction is modeled as a safety Markov decision process decision process (safety MDP). A safety MDP is a tuple \begin{small}$(\mathcal{S}, \mathcal{A}, \textrm{T}, r, \gamma, \vs_0, \omega(\vs, \va))$\end{small}, where \begin{small}$\mathcal S$\end{small} and \begin{small}$\mathcal A$\end{small} are the state and action spaces, \begin{small}$\textrm{T}(\cdot|\vs,\va)$\end{small} is the transition probabilities,
 \begin{small}$r(\vs,\va)$\end{small} is the reward function,
 \begin{small}$\gamma\in[0,1)$\end{small} is the discount factor, \begin{small}$\vs_0\in\mathcal S$\end{small} is the initial state, and 
$\omega(\vs, \va) \in \{ 0, 1\}$ is the safety violation function, that indicates whether the current state and action lead to a safety violation ($1$) or no safety violation ($0$).
Given a policy \begin{small}$\mu$\end{small}, we define the expected return as \begin{small}
$\mathcal R_\mu(\vs_0):=\mathbb E[\sum_{t=0}^{\infty}\gamma^t r(\vs_t, \va_t)\mid \mu,\vs_0]$\end{small} and at each given state $\vs\in\mathcal S$ the safety constraint function (i.e.,~expected safety violation) as \begin{small}$\mathcal W_\mu(s):=\mathbb E[\omega(\vs,\va)\mid \mu, \vs]$\end{small}. The {\em safety constraint} is then defined as \begin{small}$\mathcal W_\mu(s)\leq \epsilon$\end{small}, where $\epsilon\in[0,1]$ is the tolerable threshold of violation.
For each task the goal in safety MDP is to satisfy the safety constraint while maximizing expected return.

\begin{figure*}
    \centering
    \includegraphics[width=\textwidth,  trim={0cm 1.95cm 0cm 6.5cm}, clip]{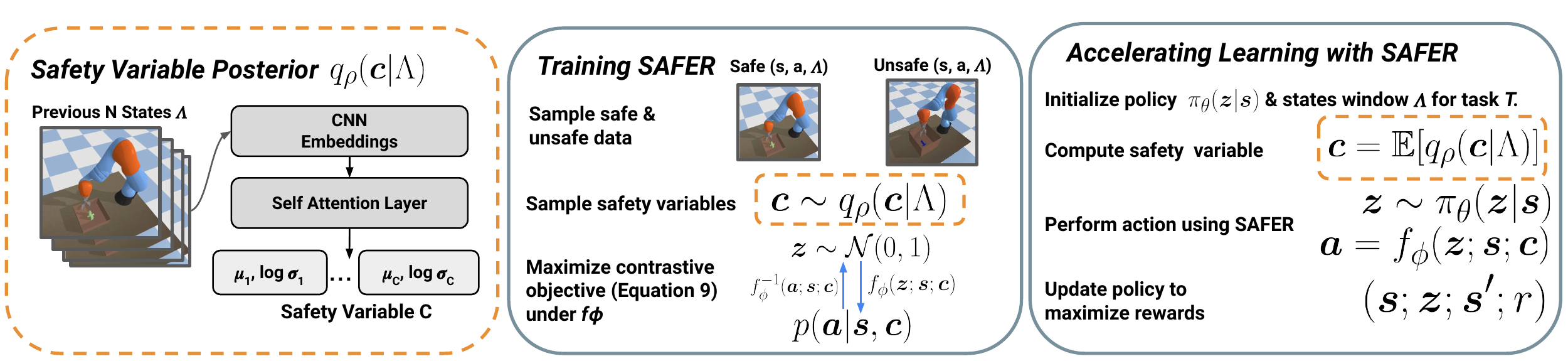}
    \caption{\textbf{Overview of SAFER:} SAFER optimizes the posterior over a latent \textit{safety variable} (left hand side of figure) that encodes safety information of the environment. SAFER  uses the safety variable to learn an abstract action space $\mathcal{Z}$ that maps to safe and useful behaviors through $f_\phi$ through a normalizing flow (middle of figure). 
    SAFER accelerates RL training by learning a latent-action policy $\pi_\theta (\vz | \vs)$ in $\mathcal{Z}$ (right hand side of figure). 
    }
    \label{fig:overview}
\end{figure*}

\textbf{Offline Primitive Learning} 
In many problems, we may not have access to the underlying reward and safety violation functions across many different tasks. Instead, we assume access to an offline dataset $\mathcal{D}$, which consists of state-action rollouts $\tau = \{\vs_0, \va_0, ..., \vs_t, \va_t \} $ collected across different tasks, where the reward and safety violation are labeled for each state action pair.
Further, when adapting to \textit{new tasks}, we assume that we \textit{do not} have access to the underlying safety violation function and only have a sparse reward signal for whether the task was completed successfully.
Thus, the safety constraints must be inferred entirely from the data.

To use the offline dataset $\mathcal{D}$ to generalize to downstream tasks, offline \emph{primitive discovery} techniques \citep{Singh2021ParrotDB, Ajay2021OPALOP, pertsch2020accelerating, pertsch2021guided}, use a policy structure consisting of a \textit{prior} $\mu_\psi=f_\phi ( \vz; \vs )$ and \textit{policy} $\vz \sim  \pi_\theta ( \vz | \vs)$.
In this parameterization, 
the prior $
    f_\phi : \mathcal{Z} \times \mathcal{S} \rightarrow \mathcal{A}
    \label{eq:mapping}$
with learnable parameters $\phi$ maps from the abstract action space $\mathcal{Z}$ and state space $\mathcal{S}$ to the action space $\mathcal{A}$ and is trained to learn a set of useful skills from the dataset $\mathcal{D}$. The task-dependent, high-level policy $\pi_\theta:\mathcal{S}\rightarrow\mathbb P(\mathcal Z)$ maps any state $\vs\in\mathcal S$ to the corresponding distribution of abstract actions in $\mathcal Z$. 
In this way, policies $\phi_\theta$ trained on downstream tasks learn to compose the primitives learned by $f_\phi$ from the offline dataset.
Different ways to express the behavior prior mapping have been proposed and have been found to greatly accelerate policy learning.
For instance, \citet{Ajay2021OPALOP} optimizes the likelihood of actions, conditioned on the state and abstract action space, $\textrm{log} \; \pi_\theta (\va | \vs, \vz) $. \citet{Singh2021ParrotDB} directly optimizes the log-likelihood, $ \textrm{log} \; p ( \va | \vs )$, and fix an invertible mapping through the use of a conditional normalizing flow \citep{realnvp} between the abstract action space $\mathcal Z$ and the distribution over useful actions $ p ( \va | \vs )$.

\textbf{Issues With Offline Primitive Discovery for Safe RL} Though current offline primitive discovery methods are highly useful at accelerating learning, they only \textit{increase} the likelihood of useful actions.
Thus, when applied to a safety MDP problem, data containing unsafe or unsuccessful data should not be used because it is counter-intuitive to increase the likelihood of these actions \citep{Singh2021ParrotDB, Ajay2021OPALOP}.
Consequently, unsafe states and actions may be out of distribution (OOD).
It is well established in the literature on deep generative models (including the techniques used in offline primitive discovery methods) that OOD data is handled poorly and, in some cases, might have higher likelihood than in-distribution data \citep{knowwhattheydontknow, classcond, whyfailood}.
As we see in Figure~\ref{fig:tradoffexample}, these observations hold true for current techniques where unsafe data has high likelihood, indicating that they may encourage unsafe behavior.
Since the proposed offline primitive discovery policy structure relies on high likelihood actions from the prior \citep{Ajay2021OPALOP, Singh2021ParrotDB}, using the aforementioned behavior priors for safety will be problematic.

\vspace{-3mm}
\section{SAFER: Safety Skill Priors}
\vspace{-2mm}

Considering the shortcomings mentioned in Section~\ref{sec:background} of existing offline primitive discovery techniques and the need for methods that can learn complex safety constraints, ideally a method that encourages safety should {\bf i)} be capable of learning complex safety constraints by sufficiently exploiting the data, thereby avoiding the OOD issue; {\bf ii)} permit the specification of undesirable behaviors through data; and {\bf iii)} accelerate the learning of successful policies. Motivated by these requirements, in this section we introduce SAFER, an offline primitive learning method that circumvents the aforementioned shortcomings and is specifically designed for safety MDPs.

\vspace{-3mm}
\subsection{Latent Safety Variable}
\label{subsec:scv}
\vspace{-2mm}

To address these criteria, we a latent variable called the \textit{safety variable} $\vc \in \mathcal{C}$ that encodes \textit{safety context} about the environment, i.e., 
$f_\phi: \mathcal{Z}\times\mathcal{C}\times \mathcal{S}\rightarrow \mathcal{A}.$
This construction encodes information beyond the current state $\vs$ to help SAFER model complex per task safety dynamics.
For example, the safety variable could encode the locations of people or animals while a robot performs household tasks.
Because we do not assume the task variable $\mathcal{C}$ is provided, we infer it from a network.

\vspace{-3mm}
\subsection{Learning The Safety Variable}
\label{sec:objective}
\vspace{-2mm}

In order to train the prior $f_\phi$ and posterior over the safety variable, we adopt a variational inference (VI) approach.
We jointly train an invertible conditional normalizing flow $f_\phi$ \citep{realnvp} as the prior $f_\phi$ and posterior over the safety variable using VI.
At each state $s\in\mathcal S$ and safety variable $c\in\mathcal C$, the flow model $f_\phi$ maps a unit Normal abstract action $z\in\mathcal{Z}$ (i.e., samples $z=f^{-1}_\phi (\va | \vs, \vc)$ of the inverse flow model follow the distribution $p_{\mathcal{Z}}(\cdot):=\mathcal N(0,I)$) onto the action space $\mathcal A$ of safe behaviors, and thus, the corresponding prior action distribution is given by 
\begin{equation}\label{eq:realnvploss}
p_\phi(\va | \vs, \vc) := p_{\mathcal{Z}}
(
f^{-1}_\phi (\va ; \vs; \vc)) \cdot | \text{det}(
\partial f^{-1}_\phi(\va ; \vs; \vc)/\partial \va) | .
\end{equation}
The flow model is a good choice for the mapping $f_\phi$ because it allows computing exact log likelihoods.
Further, it yields a mapping such that actions taken in the abstract action space $\vz\in\mathcal{Z}$ can easily be transformed into useful ones $\va = f_\phi(\vz; \vs; \vc)$.
However, since VI approximates the lower bound of maximum likelihood, it does not explicitly enforce the safety requirements in the safety variable $\vc$. To overcome this issue, we encode safety to $\vc$ by formulating the learning problem as a chance constrained optimization \citep{charnes1959chance} problem.

\textbf{Chance Constrained Optimization} Formally, our objective arises from optimizing a neural network to infer the posterior over the safety variable $\mathcal{C}$ using amortized variational inference \citep{Zhang2019AdvancesIV}.
In particular, we parameterize the posterior over the safety variable as $q_\rho ( \vc \, | \, \Lambda)$, where $\vc$ is the safety variable, and $\Lambda$ is information from which to infer the  variable. 
We set $\Lambda$ as a sliding window of states, such that if $\vs_t$ is the current state at time $t$ and $w$ is the window size, then the information is given by $\Lambda = \left[\vs_t, \vs_{t-1},...,\vs_{t-w}\right]$. 
We infer the safety variable from the sliding window of states $\Lambda$ because we expect $\Lambda$ to contain useful information concerning safe learning.
For example, in a robotics setting where the observations are images, previous states may contain useful information concerning the locations of objects to avoid, which may be unobserved in the current state.
We write the evidence-lower bound (ELBO) of our model as 
\begin{align}
    & \mathbb{E}_{\vc \sim q_{\rho} \left( \cdot | \Lambda \right)} \left[ \textrm{log} \; p_\phi (\va | \vs, \vc) \right] - D_{\textrm{KL}} (q_{\rho}(\cdot | \Lambda) || p(\cdot) ),  \label{safetynonvp}
\end{align} 
where $\va$ is the \emph{safe} action (i.e, $\omega (\va , \vs) = 0$) and $p(\vc)$ is a prior over the safety variable $\vc$.
To ensure that SAFER only samples unsafe actions with low probability, we add a chance constraint about the likelihood of unsafe actions \citep{chanceconstraints} to the ELBO optimization, 
\begin{equation}
\begin{aligned}
     \max_{\rho,\phi,\xi}\,& \mathbb{E}_{\vc \sim q_{\rho} \left( \cdot| \vs \right)} \left[  \textrm{log} \; p_\phi (\va | \vs, \vc)\right] \!-\!  D_{\textrm{KL}} \left( q_{\rho}(\cdot | \Lambda) || p(\cdot) \right) - \lambda'\xi \\ & \textrm{s.t.} \,\,\mathbb{P}_{\vc \sim q_{\rho} \left( \cdot| \vs \right)} ( p_\phi (\va_{\textrm{unsafe}} | \vs, \vc) > \epsilon) \leq \xi, \label{mixedsafety}
\end{aligned}
\end{equation}
where the constraint states that with probability $\xi$ with the safety variable $c$ drawn from $\mathcal C$ the distribution of the corresponding unsafe actions (i.e., $\omega(\va_{\textrm{unsafe}}, \vs) = 1$) is always less than the safety threshold $\epsilon$.
Intuitively, this objective enforces that the safety variable makes safe actions as likely as possible while minimizing the probability of unsafe actions. 

\textbf{Tractable Lower Bound} Due to the difficulty in optimizing the chance constrained ELBO objective, we instead consider optimizing an unconstrained surrogate lower bound~\citep{chanceconstraints}. We provide a proof in Appendix Section~\ref{ap:sec:tractablelowerbound}.
\vspace{-2mm}
\begin{proposition}
Assuming the chance constrained ELBO is written as in Equation~\ref{mixedsafety}, we can write the surrogate lower bound as,
\begin{equation}
\begin{aligned}
    \max_{\rho,\phi}\,  \mathbb{E}_{\vc \sim q_{\rho} \left( \cdot| \vs \right)} & \! \Big[  \emph{log} p_\phi (\va | \vs,\! \vc)  \!-\! \lambda  \emph{log} p_{\phi} (\va_{\textrm{unsafe}} | \vs,\! \vc)\Big] -  D_{\textrm{KL}} (q_{\rho}(\cdot | \Lambda) || p (\cdot) ) \label{objective-final} 
\end{aligned}
\end{equation}
\label{prop:1}
\end{proposition}
\vspace{-5mm}
We denote this objective as the \textit{SAFER Contrastive Objective}.
Further, this objective function has an intuitive interpretation.
The first two terms act as a contrastive loss that \emph{encourages} safe actions (high likelihood) while \emph{discourages} unsafe ones (low likelihood). 
Together with the final term, the variable $\vc$ is forced to contain useful information about safety.
Thus the objective satisfies our goals, allowing for the inference of safety constraints through the task variable and discouraging unsafe behaviors.
Finally, since SAFER can increase the likelihood of any safe behaviors, the final criteria that the offline primitive discovery technique can accelerate downstream policy learning will be met by using safe and successful trajectory data during SAFER training.

\begin{wrapfigure}{R}{0.58\textwidth}
    \begin{minipage}{0.578\textwidth}
  \begin{algorithm}[H]
    \caption{Accelerating Safe Reinforcement Learning with SAFER}
    \label{alg:usingsafer}
    \begin{algorithmic}
      \REQUIRE SAFER Prior $f_\phi$, Safety Posterior $q_\rho ( \vc | \Lambda )$, 
      Safety bound $\eta$, Task $\mathcal{T}$, Window $\Lambda  = \{\}$
      \FOR {step $k = 1,...,K$}
        \STATE $\vs_k \leftarrow$ current state
        \STATE $\vc_k \leftarrow \mathbb{E}_{\vc\sim q_\rho (\cdot | \Lambda_k)} \left[ \vc \right]$\hfill \algorithmiccomment{{ \color{magenta}\texttt{Mean safety var.}}}
        \STATE $\vz_k \sim \pi_\theta \left( \cdot | \vs_k \right)$ \hfill\algorithmiccomment{{\color{magenta}\texttt{Sample abstract action}}}
        \STATE $\va_k\leftarrow f_\phi ( \vz_k ; \vs_k ; \vc_k )$ \hfill\algorithmiccomment{{\color{magenta}\texttt{Get SAFER action}}}
        \STATE $s_{k+1}, r_k, \omega_k \leftarrow$ Perform $\va_k$ in task $\mathcal{T}$
        \STATE Update $\pi_\theta(\vz | \vs)$ using $(\vs_k, \vz_k, \vs_{k+1}, r_k)$
        \STATE Update $\Lambda$ with $\vs_k$ in FIFO order 
      \ENDFOR
      \STATE \textbf{Return:} Policy $\pi_\theta(\vz | \vs) $ for task $\mathcal{T}$
    \end{algorithmic}
  \end{algorithm}
  \end{minipage}
\end{wrapfigure}

\textbf{Parametization Choices} To parameterize the SAFER action mapping $f_\phi$, we use the Real NVP conditional normalizing flow, proposed by \citet{realnvp}, due to it being highly expressive and allowing exact log-likelihood calculations.
Next, we parameterize the posterior distribution $q_\rho (c | \Lambda)$ over the safety variable as a diagonal Gaussian to compute the KL efficiently while enabling an expressive latent space.
We use a transformer architecture to model the sequential dependency between Gaussian safety  variable $\vc$ and the window of previous states $\Lambda$ \citep{NIPS2017_3f5ee243}.
Finally, because the state space is an image pixel space, we also encode each observation to a vector using a CNN.
An overview of the architecture is given in Figure~\ref{fig:overview}.

\textbf{Training} It is necessary to use the reparameterization trick to compute gradients across the objective in Equation~\ref{objective-final} \citep{Kingma2014AutoEncodingVB}. 
Second, optimizing Equation~\ref{objective-final} involves minimizing an unbounded log-likelihood in the second term of the objective.
This term can lead to numerical instabilities when $p_\phi ( \va_{\textrm{unsafe}} | \vs, \vc )$ is too small.
To overcome these issues, we use gradient clipping and freeze this term if it starts to diverge. Psuedo code of the procedure to train SAFER is provided in Appendix~\ref{ap:safer} in Algorithm~\ref{alg:trainingsafer} and hyperparameter details are provided in Appendix~\ref{ap:hyperparameterdetails}.
\vspace{-3mm}
\subsection{Accelerating Safe RL with SAFER}
\vspace{-2mm}
\nevan{I think you should mention earlier that the RL agent will generate an action in the abstract action space and the flow model will be used to compute the real action. This could go in the introduction.}\dylan{added}
When using SAFER on a safe RL task, the goal is to accelerate safe learning by leveraging the mapping $f_\phi$ in the hierarchical policy $\mu_\psi(\vs,\vc) = \int_\vz f_\phi (\vz ; \vs ; \vc ) d\pi_\theta ( \vz | \vs)$ where the policy parameters of the mapping $\phi$ are fixed and the parameters $\theta$ need to be optimized (Psuedo code of the procedure is provided in Algorithm~\ref{alg:usingsafer}).
The policy $\pi_\theta(\vz | \vs)$ can be learned by any standard RL methods (e.g., SAC~\citep{Haarnoja2018SoftAO}) that produces continuous actions.
To leverage SAFER at inference time, at each timestep $t$ the RL policy takes an action in the abstract action space $\vz_t\sim\pi_\theta ( \vz | \vs = \vs_t)$.
Using the sliding window of states $\Lambda$, the safety variable posterior computes the distribution over the safety variable $c_t$.\footnote{If there are insufficient states to compute a task window of size $w$ (e.g., at the beginning of the rollout), we pad the available states with $0$'s in order to construct a window of $w$ states.}
Because a single safety variable value $c_t$ is required, we fix it at its mean, $\textrm{E}[ \vc_t ] = \int \vc \, dq_\rho (\vc | \Lambda_t) $.
Finally, SAFER computes the action $\va_t = f_\phi (\vz_t ; \vs_t ; \textrm{E}[ \vc_t ])$, the action is taken the environment, and the reward $r(\vs_t, \va_t)$  and safety violations $\omega(\vs_t, \va_t)$ are returned.
The action $\vz_t$ and reward $r_t$ are added to the replay buffer for subsequent RL training.

\vspace{-4mm}
\subsection{Using SAFER to Guarantee Safety}
\label{sec:assurances}
\vspace{-2mm}
Next, we demonstrate how it is straightforward to use SAFER to theoretically guarantee safety for \textit{any} policy trained under the prior.
To show this is the case, we assume there always exists safe actions to take in the environment and make an optimiality assumptions about the prior in~(\ref{objective-final}).
Then, we can construct a bound on the range of abstracts actions that ensures only safe actions under the prior:
\nevan{I think the proposition should have more justification.}\dylan{I'll point readers to the proof in the appendix}
\vspace{-2mm}
\begin{proposition}
There exists an $\eta$ such that the corresponding bounded abstract actions $\vz \in (-\eta, \eta)$ are safe, i.e., $\omega ( \vs, f_\phi ( \vz ; \vc ; \vs )) = 0 $, $\forall \vz \in (-\eta, \eta), \vs, \vc$.
\label{prop:optimaleta}
\end{proposition}
\vspace{-2mm}
As a result, we can construct a latent variable bound around the mean of $\mathcal{Z}$ as the actions $f_\phi ( \vz ; \vc ; \vs )$ that are more likely to be safe and successful are closer to the mean. 
Because unsafe actions under the SAFER prior have lower likelihood and our assumption ensures that safe actions exists, there must exist a finite latent variable bound $\eta$ that contains all safe actions.
Consequently, with such an $\eta$ from Proposition~\ref{prop:optimaleta}, any agent $\pi_\theta$ that is trained under the SAFER prior and has a bounded abstraction action output $\vz \in (-\eta, \eta)$ is safe.
The full proof details are provided in Appendix~\ref{ap:gaurantees}.

In practice, we use an offline data set with safe ($\vs$, $\va$) and unsafe ($\vs$, $\va_{\text{unsafe}}$) state-action pairs to determine the value of $\eta$ that ensures safety. Also, it is acceptable to fix a range ($-\eta$, $\eta$) that includes a small number of unsafe actions (e.g., at most $1-\epsilon$ portion of all actions in data is unsafe) to avoid an overly tight bound. 
We optimize the real-valued $\eta>0$ by a numerical gradient-free approach. 
First, we initialize $\eta=\eta_0$ with a large constant and use that to generate the corresponding SAFER abstract actions $\vz$ w.r.t. the offline data, whose latent variable is bounded in $(-\eta_0, \eta_0)$. We sub-sample this SAFER-action bootstrapped dataset to construct a refined one that only has at most $1-\epsilon$ portion of unsafe actions. Since the normalizing flow in SAFER is invertible, for every ($\vs$, $\va$)-pair in this data computing the corresponding latent action value is straightforward. This allows us to estimate $\eta_1$, which is the maximum latent action in this dataset. We repeat this procedure until convergence. If the offline dataset contains sufficiently diverse state-action data that covers most situations encountered by SAFER, we expect the above safety threshold to be generalizable \citep{kaariainen2005comparison}.
We denote this procedure computing the SAFER \textit{safety assurances} and provide pseudo code in Algorithm~\ref{alg:safeassure} in the Appendix.

\begin{SCfigure}
    \centering
    \includegraphics[width=.6\columnwidth,  trim={1cm 1.65cm 8.0cm 5cm}, clip]{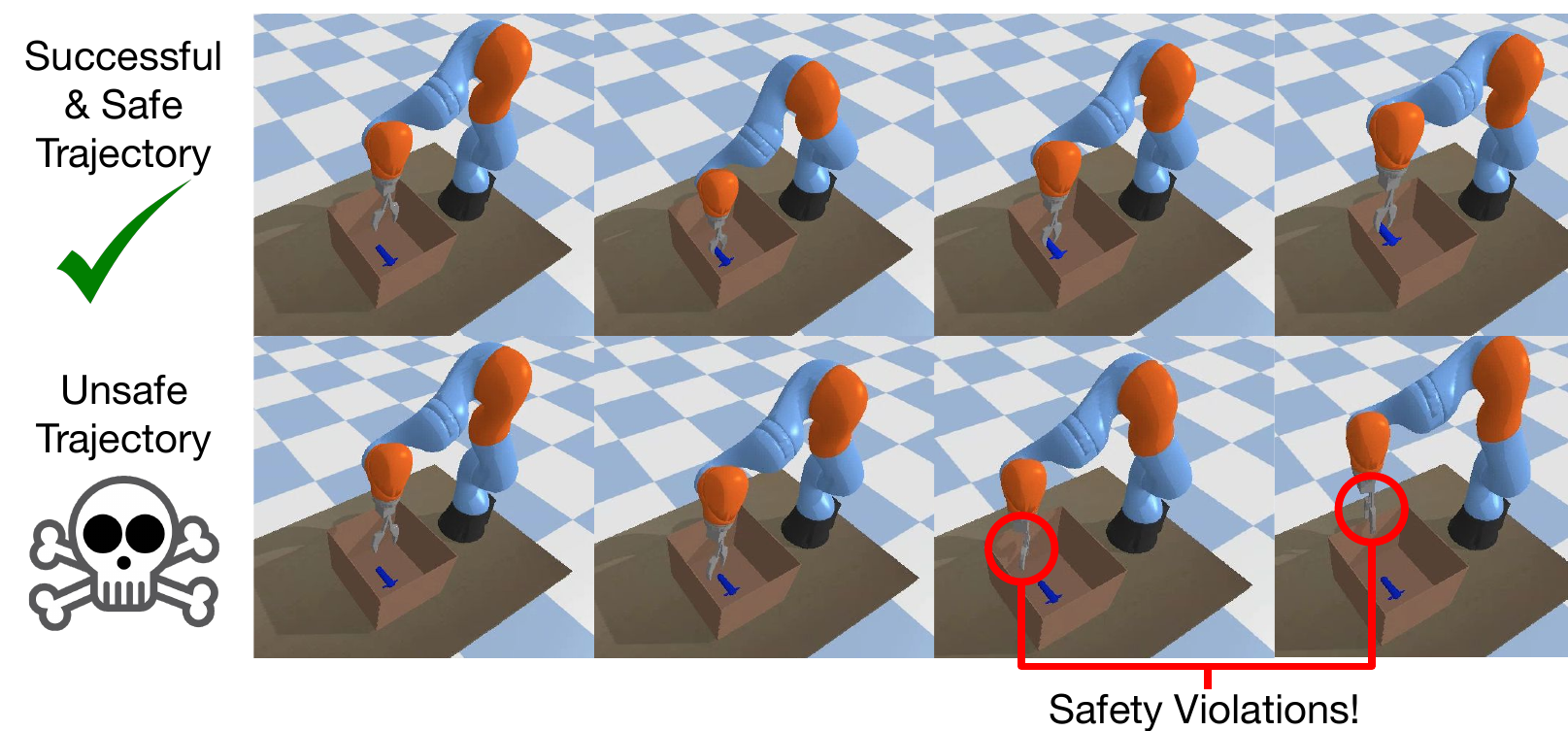}
    \caption{\textbf{An example of a task} where the robot successfully and safely grasping an object (top row). Here, the robot reaches into the container and extracts the object \textit{without} touching the container.  On the bottom row, the robot performs the same task but commits safety violations by touching the container.}
    \label{fig:tasks}
\end{SCfigure}

\vspace{-4mm}
\section{Experiments}
\vspace{-2mm}

We evaluate the calibration of the safety assurances introduced in Section~\ref{sec:assurances} and how well SAFER encourages both safe and successful policy learning compared to baselines.
\vspace{-3mm}
\subsection{Experiments Setup}
\label{subsec:setup}
\vspace{-2mm}

To evaluate SAFER, we introduce a suite of safety-critical robotic grasping tasks that are inspired by the game Operation\footnote{\href{https://en.wikipedia.org/wiki/Operation_(game)}{\texttt{https://en.wikipedia.org/wiki/OperationGame}}}.

\textbf {Safety-critical Robotic Grasping Tasks}
Based on the game Operation, whose goal is to extract objects from different sized containers without touching the container, we construct a set of $40$ grasping tasks, each consisting of a container and object defined in PyBullet \citep{coumans2021}. 
We collect data from these tasks to train SAFER and use $6$ of the more complex tasks for evaluation.
In our tasks, the objects are randomly selected from ones available in PyBullet package, and the containers are generated to fit the objects, whose dimensions (heights and widths) are generated randomly. Our agent controls a $5$DoF robotic arm and gripper.
The agent receives positive reward ($r(s,a)=1$) when it extracts the object from the box and a negative reward ($r(s,a)=-1$) at every time step while the task is incomplete. The agent incurs a safety violation ($\omega(s,a)=1$) if the arm touches the box (examples of safe/unsafe trajectories in Figure~\ref{fig:tasks}, examples of the tasks are in Figure~\ref{fig:add_task_examples}). The states are $48\times48$ pixel image observations of the scene collected from a fixed camera. 

\textbf{Offline Data Collection}
To generate the offline data for the SAFER training algorithm, for each robot grasping task we use the scripted policy from \citet{Singh2021ParrotDB} to collect trajectories with a total of $1,000,000$ steps. The scripted policy controls the robotic arm to grasp the object generally by minimizing the absolute distance between objects and the robot. To obtain more diverse/exploratory trajectories, one also adds random actuation noise to the policy.
After collecting the trajectories, for each state-action pair $(\vs, \va)$ in the dataset we provide labels for \textbf{i)} safety violation $\omega(s,a)\in\{0,1\}$, and \textbf{ii)} whether the pair $(\vs, \va)$ is part of a successful rollout (i.e., $(\vs, \va)$ such that $\mathbb E[r(\vs_T,\va_T)|\mu_{\text{data}},\vs_0=\vs,\va_0=\va]=1$, where $T$ is the trajectory length random variable). To create the state window $\Lambda$ for SAFER training, for each $(\vs, \va)$ in the data buffer we save the previous $w$ states. One can utilize these labels to categorize safe versus unsafe data to train SAFER.

\textbf{Baseline Comparisons} To demonstrate the improved safety performance of SAFER over existing offline primitive learning techniques, we compare against baseline methods that leverage offline data to accelerate learning, including PARROT \citep{Singh2021ParrotDB}, a contextual version of PARROT (Context. PAR) that uses a latent variable to help accelerate learning, Prior Explore (a method that samples from SAFER to help with data collection during training) and RL from scratch using SAC. See Appendix~\ref{sec:baselines} for more details.
Last, because our setting requires learning safety constraints \textit{entirely} from labeled offline data, we do not compare against methods that require online safety constraint functions, such as many existing safe RL methods which use a constraint function during training.

\vspace{-2mm}
\subsection{Results Discussion}
\label{subsec:rl}
\vspace{-1mm}

\begin{table*}[t]
\small
\centering
\caption{\textbf{Training RL with SAFER}, we give the mean $\pm$ SD success rate and cumulative safety violations across different tasks and initializations. SAFER produces the lowest cumulative safety violations throughout training and outperforms the baseline methods in terms of success rate. 
} 
\label{tab:rlresults}
\resizebox{.95\textwidth}{!}{%
\begin{tabular}{l rrrrrr rrrrrr} 
& \multicolumn{6}{c}{\bf Success Rate (\%)} & 
\\
 \cmidrule(lr){2-7}
&  \multicolumn{1}{c}{\bf Task 1} & \multicolumn{1}{c}{\bf Task 2} & \multicolumn{1}{c}{\bf Task 3} &  \multicolumn{1}{c}{\bf Task 4} &
\multicolumn{1}{c}{\bf Task 5} &
\multicolumn{1}{c}{\bf Task 6} & \\
\midrule 
SAC & $0.0\pm0.0$ &  $0.0\pm0.0$  & $0.0\pm0.0$  & $0.0\pm0.0$ & $0.0\pm0.0$  &  $2.3\pm0.0$   \\  
PARROT   & $0.0\pm0.0$ &  $12.8\pm0.2$  & $25.7\pm0.2$  & $16.1\pm0.2$ & $33.9\pm0.3$  &  $6.3\pm0.1$   \\
Context PAR.  & $5.0\pm0.0$ &  $24.2\pm0.2$  & $27.0\pm0.3$  & $0.7\pm0.0$ & $7.3\pm0.1$  &  $12.0\pm0.2$   \\
Prior Explore  & $1.8\pm0.0$ &  $1.5\pm0.0$  & $3.0\pm0.0$  & $1.8\pm0.0$ & $1.1\pm0.0$  &  $1.0\pm0.0$   \\ 
SAFER & $\boldsymbol{ 21.0\pm0.1 } $ &  $\boldsymbol{ 87.4\pm0.2 }$  & $\boldsymbol{ 89.3\pm0.0 }$  & $\boldsymbol{ 28.1\pm0.2 }$ & $\boldsymbol{ 54.4\pm0.1 }$  &  $\boldsymbol{ 83.3\pm0.0 }$   \\ 

 \rule{0pt}{3ex}  &  \multicolumn{6}{c}{\bf Total Number of Safety Violations (Out of $\textbf{50,000}$ Steps)} & 
\\
 \cmidrule(lr){2-7}
&  \multicolumn{1}{c}{\bf Task 1} & \multicolumn{1}{c}{\bf Task 2} & \multicolumn{1}{c}{\bf Task 3} &  \multicolumn{1}{c}{\bf Task 4} &
\multicolumn{1}{c}{\bf Task 5} &
\multicolumn{1}{c}{\bf Task 6} & \\
\midrule
SAC & $2045\pm236$ &  $876\pm117$  & $1055\pm216$  & $2736\pm147$ & $2188\pm405$  &  $756\pm293$   \\  
PARROT   & $6332\pm3026$ &  $307\pm291$  & $13\pm21$  & $541\pm461$ & $2414\pm314$  &  $932\pm844$   \\
Context PAR.  & $5929\pm2964$ &  $1576\pm1208$  & $1039\pm777$  & $5056\pm1778$ & $2796\pm624$  &  $2085\pm1951$   \\
Prior Explore  & $6203\pm551$ &  $2240\pm634$  & $2867\pm853$  & $4525\pm826$ & $4669\pm542$  &  $2596\pm703$   \\ 
SAFER & $\boldsymbol{610\pm184}$ &  $\boldsymbol{51\pm61}$  & $\boldsymbol{10\pm14}$  & $\boldsymbol{455\pm470}$ & $\boldsymbol{1707\pm292}$  &  $\boldsymbol{7\pm9}$   \\
\bottomrule
\end{tabular}
}
\end{table*}%
\textbf{Effectiveness of RL training with SAFER} In Table~\ref{tab:rlresults} we compare SAFER with the baseline methods both in terms of cumulative safety violations and success rate.
Note, here we use the underlying reward and safety violation functions for each task to evaluate performance.
We choose a SAFER policy primitive with a safety assurance upper bound that guarantees at most $15\%$ unsafe actions, which empirically maintains a good balance between performance and safety.
For each downstream task, we then train the RL agent $\pi_\theta$ with SAC for only $50,000$ steps because we are more interested to evaluate the power of the primitive learning algorithm. 
Overall, we see that SAFER has the lowest cumulative safety violations, indicating that it is the most effective method in promoting safe policy learning.
Interestingly, SAFER also consistently outperforms other methods in policy performance.
The strong success rates of SAFER are potentially due to the fact that discouraging unsafe behaviors may indeed help refining the space of useful behaviors, thus improves policy learning.

\begin{figure*}
    \centering
    \includegraphics[width=.8\textwidth, trim={1.25cm 1.25cm 1.25cm 0},clip]{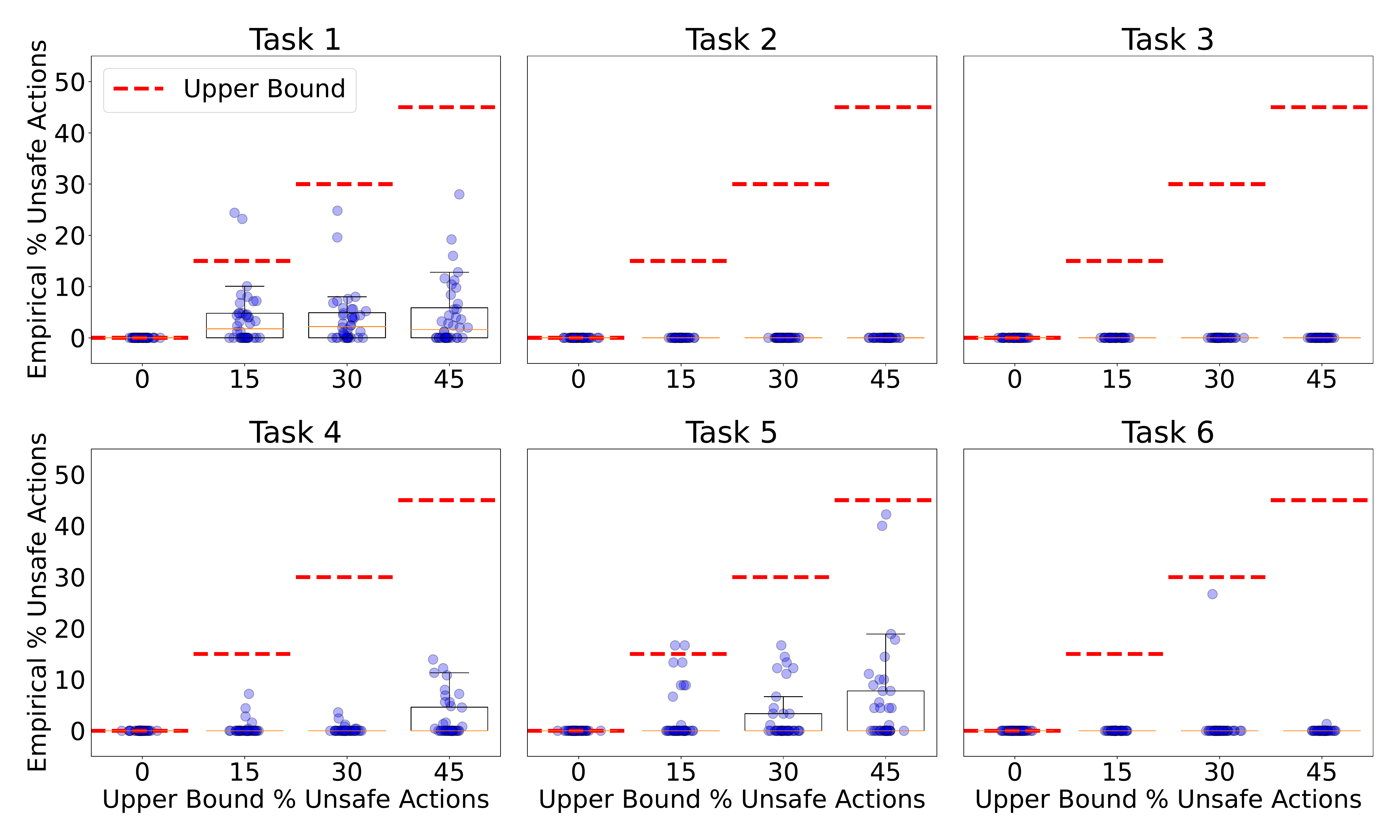}
    \caption{\textbf{Assessing the calibration of the SAFER safety assurances} by randomly sampling actions from the prior with various safety upper bounds across different evaluation tasks. Each dot corresponds to the empirical percent of unsafe $(\vs, \va)$ pairs from a single rollout on the task. Overall, we see that the SAFER safety assurances are quite well calibrated.}
    \label{fig:calibration}
\end{figure*}
\textbf{Safety Assurance Calibration} We evaluate whether the safe abstract action bound of SAFER computed in Section~\ref{sec:assurances} is well calibrated, i.e., the empirical percent of unsafe actions should be less than the upper bound.
To study this, we compute the $\mathcal Z$-action bound $(-\eta, \eta)$ corresponding to an upper bound of $0\%$, $15\%$, $30\%$ and $45\%$ unsafe actions for SAFER.
We compute the percentage of unsafe actions by randomly sampling actions from SAFER on each evaluation task and report the results in Figure~\ref{fig:calibration}, showing that the SAFER bounds are indeed well calibrated. 

\textbf{Impact of latent safety variable} %
We train SAFER on Tasks $2$ and $5$ using the contrastive objective in Equation~\ref{objective-final} but without the safety variable.
In this case, the success rate never exceeds $10\%$ and the safety violations are quite high (see Appendix~\ref{ap:generalization} for the Task 2 results). In contrast, the safety variable in SAFER has at least a $60\%$ success rate on both tasks ( Table~\ref{tab:rlresults}). This result suggests that the latent safety variable is crucial for success and safety.

\vspace{-3mm}
\section{Limitations}
\vspace{-2mm}

Though SAFER improves both safe and successful generalization to downstream tasks, there are several critical limitations to consider.
Foremost, SAFER relies on a labeled offline dataset.
In certain settings, it may be impractical to collect a sufficiently large dataset to ensure useful learned primitives or to receive high quality labels.
If the dataset is not sufficiently large or the labels are poor quality, this could harm the capacity of SAFER to learn useful primitives.
In the future, researchers should benchmark and improve the sample efficiency of SAFER.
Second, the offline dataset includes a selection of unsafe demonstrations.
In settings where there are not existing unsafe data points (e.g., from previous failures), or unsafe data cannot be simulated, it may be difficult for SAFER to learn generalizable safety constraints from the data.

\vspace{-3mm}
\section{Conclusion}
\vspace{-2mm}

In this paper, we introduced SAFER, an offline primitive learning method that improves the data efficiency of safe RL when there is access to both safe and unsafe data examples.
This is particularly important because most existing safe RL algorithms are very data inefficient.
We proposed a set of complex safety-critical robotic grasping tasks to evaluate SAFER, investigated limitations of state-of-the-art offline primitive learning baselines, and demonstrated that SAFER can achieve better success rates while enforcing safety with high-probability assurances.

\bibliography{example}

\newpage
\appendix
\onecolumn

\begin{center}
{\huge Appendix}
\end{center}

\section{Proof: Tractable Lower Bound}
\label{ap:sec:tractablelowerbound}

\textit{Proposition 3.1:} Assuming the chance constrained ELBO is written as in Equation~\ref{mixedsafety}, we can write the surrogate lower bound as,
\begin{equation}
\begin{aligned}
    \max_{\rho,\phi}\,  \mathbb{E}_{\vc \sim q_{\rho} \left( \cdot| \vs \right)} & \! \Big[  \emph{log} p_\phi (\va | \vs,\! \vc)  \!-\! \lambda  \emph{log} p_{\phi} (\va_{\textrm{unsafe}} | \vs,\! \vc)\Big] -  D_{\textrm{KL}} (q_{\rho}(\cdot | \Lambda) || p (\cdot) ) \label{objective-final} 
\end{aligned}
\end{equation}
\label{prop:1}

\textbf{Proof:}
We rewrite the optimization~\ref{mixedsafety} into the following form, 
\begin{equation}
\begin{aligned}
     \max_{\rho,\phi, \lambda'}\mathbb{E}_{\vc \sim q_{\rho} \left( \cdot| \vs \right)} & \!\left[  \textrm{log} \; p_\phi (\va | \vs, \vc) \!-\!  D_{\textrm{KL}} \!\left( q_{\rho}(\cdot | \Lambda) || p(\cdot) \right) \right] - \lambda'\mathbb{P}_{\vc \sim q_{\rho} \left( \cdot| \vs \right)} ( p_\phi (\va_{\textrm{unsafe}} | \vs, \vc) > \epsilon).
\end{aligned}
\end{equation}
With the Markov inequality we have
\begin{align}
   \mathbb{P}_{\vc \sim q_{\rho} \left( \cdot| \vs \right)} \!( p_\phi (\va_{\textrm{unsafe}} | \vs,\! \vc) \!>\! \epsilon) \!\leq\!  \frac{\mathbb{E}_{\vc}\! \left[  p_\phi (\va_{\textrm{unsafe}} | \vs,\! \vc) \right]}{\epsilon},
\end{align}
such that the following objective function is a lower bound of that in Equation~\ref{mixedsafety}:
\begin{equation}
\begin{aligned}
  \max_{\rho,\phi, \lambda'}  \mathbb{E}_{\vc \sim q_{\rho} \left( \cdot| \vs \right)} & \Big[  \textrm{log}  p_\phi  (\va | \vs, \vc)  \!-\! \frac{\lambda'}{\epsilon}  p_\phi (\va_{\textrm{unsafe}} | \vs, \vc) \Big] -  D_{\textrm{KL}} (q_{\rho}(\cdot | \Lambda) || p (\cdot))  
\end{aligned}
\end{equation}
For convenience, we write $\frac{\lambda'}{\epsilon}$ as the single hyperparameter $\lambda$ and optimize the log of $p_{\phi} (\va_{\textrm{unsafe}} | \vs, \vc)$ for better numerical stability. We finally have the lower bound surrogate objective in Equation~\ref{objective-final}.

\section{Guaranteeing Safety with SAFER}
\label{ap:gaurantees}

In this section, we demonstrate how it is straightforward to show SAFER can guarantee safety for any policy trained under the skill prior, demonstrating the utility of the method.
We restate and clarify our assumptions before providing the proof of the proposition.
The first assumption ensures that there is always a safe action to take.
\begin{assumption}
At every state $\vs$, there always exists a safe action $\va$, i.e., $\forall \vs \; \exists \va \; s.t. \; \omega(\vs, \va) = 0$.
\label{assum:first}
\end{assumption}
The second assumption ensures that the SAFER model is optimal according to the SAFER objective given in Objective~\ref{objective-final}.
In effect, this assumption means that safe actions have high likelihood while the unsafe actions are much less likely under the SAFER prior.
\begin{assumption}
The SAFER prior parameters $\hat{\rho}, \hat{\phi}$ are optimal per Objective~\ref{objective-final}, such that all safe actions have higher likelihood than unsafe actions under the prior, i.e., $\forall \va, \va_{unsafe}, \vs, \vc : \; \textrm{log} p_{\phi} (\va | \vs, \vc) \gg \textrm{log} p_{\phi} (\va_{unsafe} | \vs, \vc)$.
\label{assum:second}
\end{assumption}
Next, we provide a proof for Proposition~\ref{prop:optimaleta}.
\begin{proposition}
There exists an $\eta$ such that the corresponding bounded abstract actions $\vz \in (-\eta, \eta)$ are safe, i.e., $\omega ( \vs, f_\phi ( \vz ; \vc ; \vs )) = 0 $, $\forall \vz \in (-\eta, \eta), \vs, \vc$.
\end{proposition}
\textit{Proof (Sketch):} 
Because the abstract action space $\mathcal{Z}$ is unit Gaussian ($\mathcal{Z}\sim\mathcal{N}(0,I)$) it is the case that the $\vz$'s that are closer to the zero vector $\boldsymbol 0$ have higher likelihood, i.e., $\vz$'s with lower norm $||\vz||$ have higher likelihood. 
From the assumption, we know that in every state there exists safe actions and with the way we train SAFER it will have much higher likelihood than unsafe actions in the prior distribution, $\textrm{log} p_{\phi} (\va | \vs, \vc) \gg \textrm{log} p_{\phi} (\va_{unsafe} | \vs, \vc)$.
Using the invertibility property of normalizing flow, one concludes that for all state and action pairs, the unsafe abstract actions $\vz$ are much farther away from the zero vector $\boldsymbol 0$.
Consequently, there must exist a finite latent bound $\eta$ that separates all safe actions with unsafe ones.\hfill$\qed$

\section{Additional Results}
\label{ap:generalization}

In this Appendix, we present additional results with SAFER.

\begin{figure}[H]
    \centering
    \includegraphics[width=.8\textwidth, clip, trim={0cm 25cm 0 0}]{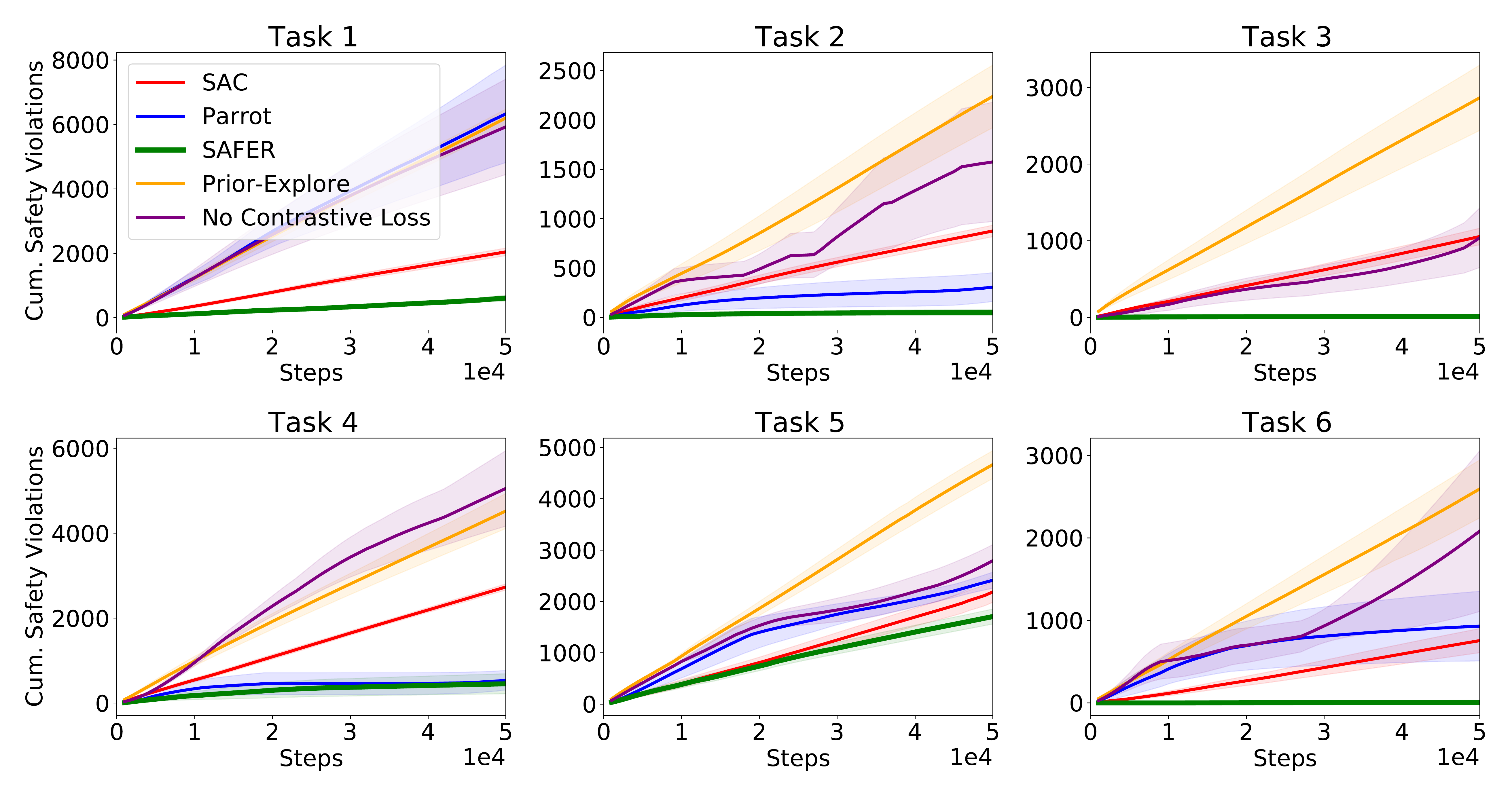}
    \caption{\textbf{The cumulative safety violations throughout training} for SAFER and the baselines. We see that SAFER is consistently the safest method throughout training.}
    \label{fig:cumsum}
\end{figure}

\textbf{Cumulative Safety Violation Graphs} In the main paper, we presented the cumulative safety violations at the end of training. Here, we present graphs of the cumulative safety violations in figure \ref{fig:cumsum} throughout training for the baselines and SAFER.
In these graphs, we see that SAFER is consistently the safety method throughout training.

\begin{figure}[t]
    \centering
    \includegraphics[width=.7\columnwidth, trim={0 1.5cm 0 0 },clip]{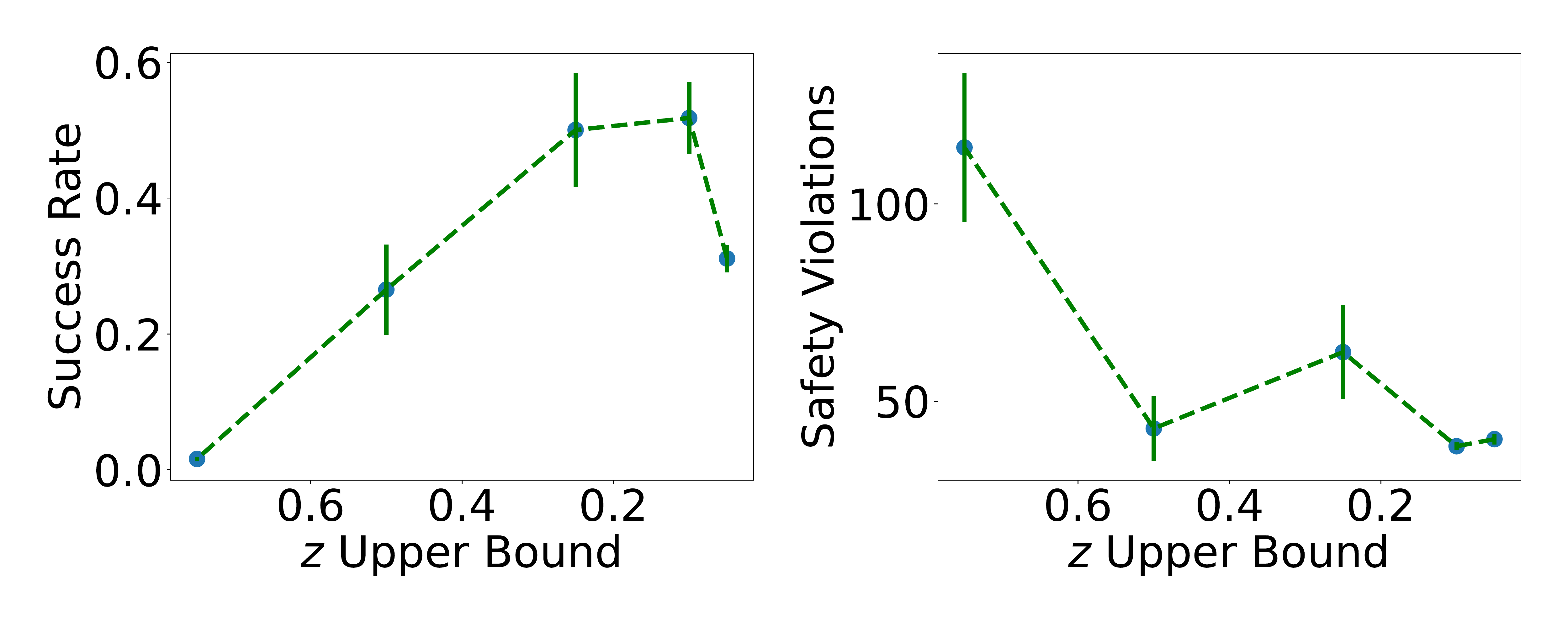}
    \caption{\textbf{Assessing the tradeoff between success and safety} varying the safety assurances bound on the abstract action space $\mathcal Z$, (referred to as $\eta$ in Algorithm~\ref{alg:safeassure}). There is an sweet spot where success rate is high and safety violations is low. 
    }
    \label{fig:successsafety}
\end{figure}

\textbf{Success \& Safety Tradeoff} In Figure~\ref{fig:successsafety} we assess the tradeoff between success and safety by varying the $\mathcal Z$-action bound in Algorithm~\ref{alg:safeassure}. We sweep over different bounds and compute both the success rate and safety violations at the end of training for Task 5.
We see that there is a sweet spot with high  success rate and low safety violations when the safety assurances bound is close to $15\%$. Interestingly when the bound is too tight (corresponding small $\vz$ values), both the safety violation and success rate become low, indicating SAFER cannot solve the task without sufficient exploration.

\begin{figure}
    \centering
    \includegraphics[width=\textwidth, clip, trim={0 28cm 0 13cm}]{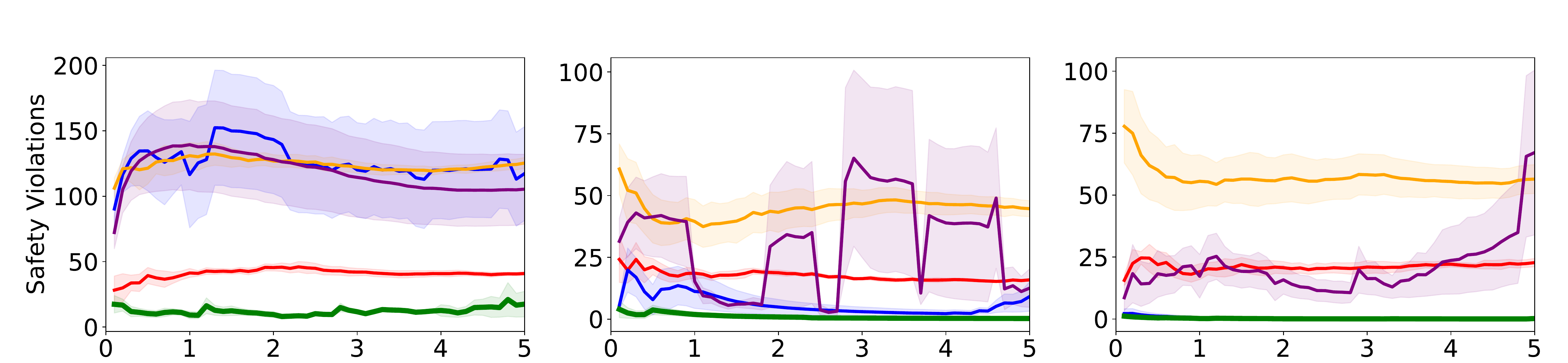}
     \includegraphics[width=\textwidth, clip, trim={0 0cm 0 40cm}]{images/resultsf2.pdf}
    \caption{The safety violations over each step of training for each of the tasks (same task ordering as Figure~\ref{fig:cumsum}). We see that SAFER is consistently the most safe method throughout training.}
    \label{fig:perstepsafety}
\end{figure}

\textbf{Per Step Safety Violations} In the main paper, we provide \textit{cumulative} safety violation graphs.
Here, we provide the safety violations over the last $1,000$ steps in Figure~\ref{fig:perstepsafety} in order to get a better sense of the safety violations throughout training.
We again consistently see SAFER is safety method over the course of training.
One interesting observation is that, in Section~\ref{sec:background}, we discussed how PARROT rates unsafe ($\vs, \va$) pairs as high likelihood.
Because PARROT draws on higher likelihood actions from the prior earlier in training, we would expect that PARROT would be more unsafe earlier in training.
Empirically, we see this to be the case.
Looking at the graphs, PARROT has high safety violation spikes at the beginning of training.
These results demonstrate that our earlier observations surrounding the unsafety of PARROT hold true when running RL.

\textbf{Impact of Probabilistic Treatment} One question worth considering is how necessary is it to treat SAFER as a latent variable model and optimize the posterior over the safety variable using variational inference, as is proposed in Section~\ref{sec:objective}.
It could be easier to treat $\vc$ as a vector (without defining it as a Guassian random variable), exclude the KL term from Equation~\ref{objective-final}, and optimize $q_\rho ( \vc | \Lambda )$ with the rest of the objective. 
To assess whether this is the case, we ran a sweep across different hyperparameter configurations, including the number of bijectors in the real NVP model, the learning rate, $\lambda$, and the number of hidden units in each bijector.
Doing this, however, we find SAFER quickly diverges, indicating the probabilistic treatment greatly helps stabilize training and is necessary for the success of the method.

\begin{figure}[t]
    \centering
    \includegraphics[width=.5\textwidth]{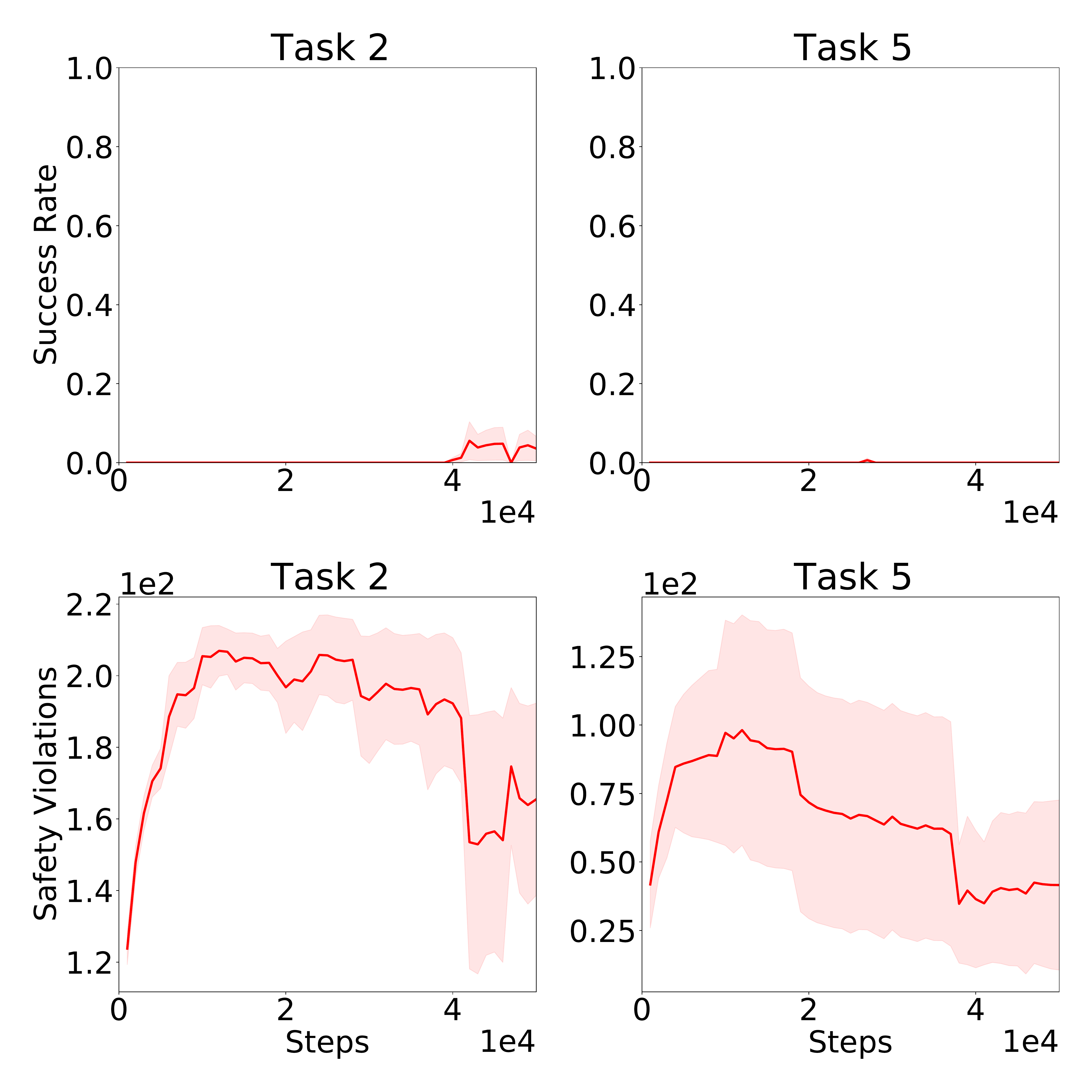}
    \caption{Effectiveness of RL Training using the SAFER objective \textit{without} the safety variable. We see the prior without the safety variable is quite unsuccessful, indicating that the safety variable is critical to enabling SAFER to promote both safe and successful learning.}
    \label{fig:novar}
\end{figure}

\textbf{Training SAFER Without the Safety Context Variable} As an abalation in the main paper, we considered training SAFER without the SAFETY context variable and found that it led to worse success rate and relatively higher safety violations.
In this Appendix, we provide the full training results in Figure~\ref{fig:novar} in terms of success rate and per step safety violations.
Here, we see that for the tasks considered, training SAFER without the safety variable leads to worse success rates and less safety (compared to the per step success rates in Figure~\ref{fig:perstepsafety}).

\begin{figure}[h!]
    \centering
    \includegraphics[width=.5\textwidth]{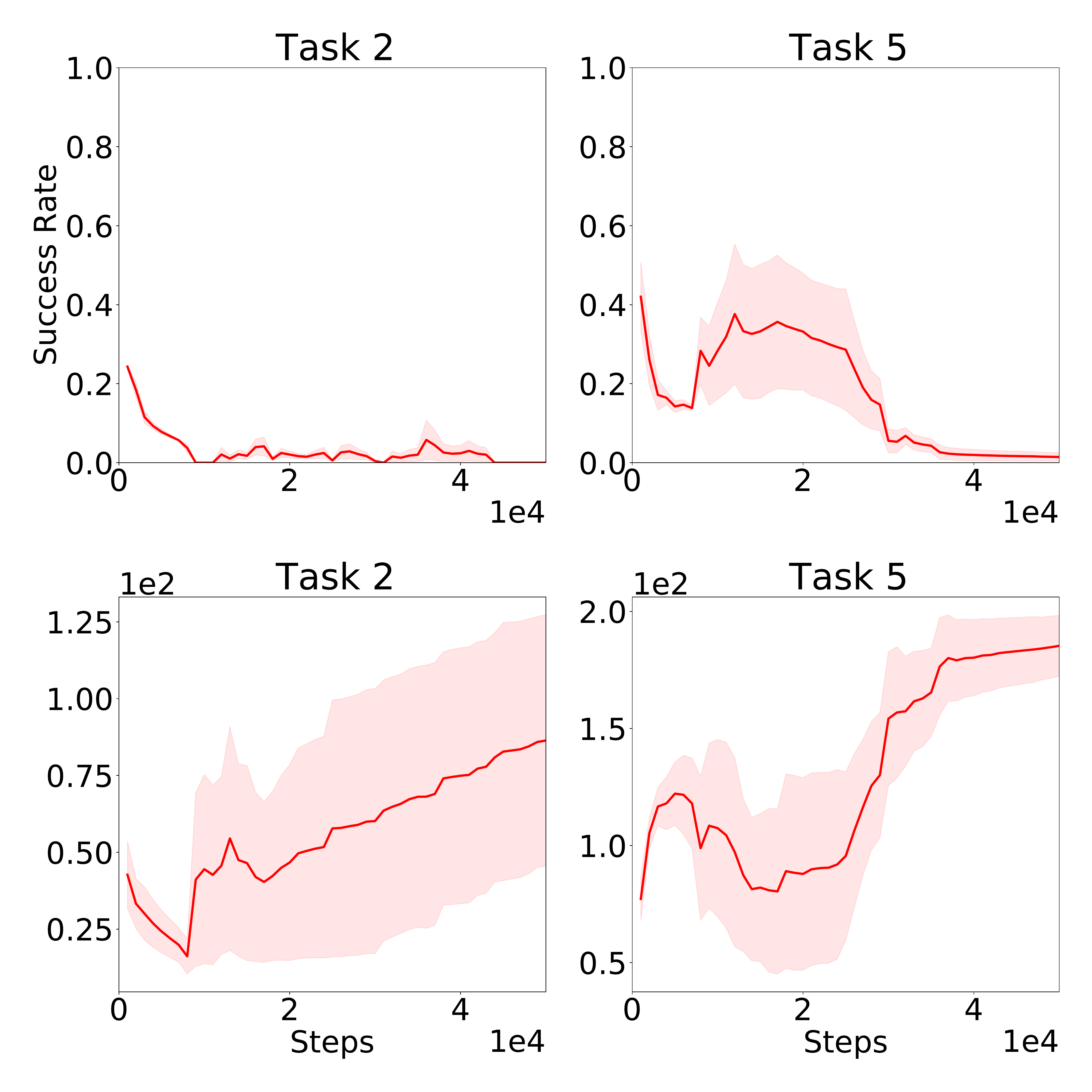}
    \caption{\textbf{Training PARROT using unsafe data} from successful trajectories as well as safe data. We see that this leads to leads to relatively worse success rates (top row) as well as relatively higher per step safety violations (bottom row). These results suggest it is best to train PARROT with safe and successful data only.}
    \label{fig:unsafedata}
\end{figure}

\textbf{Training PARROT With Unsafe Data} In the paper, we performed experiments PARROT trained using safe data. Meaning, $w(\vs, \va)=0$ for each training point. We also limited the data to only those tuples in successful trajectories to promote PARROT acquiring safe and successful behaviors.
Though it makes the most sense to train PARROT for safety concerned tasks in this fashion, it is worth considering what would happen if we also included unsafe data from successful trajectories.
To assess what would happen, we train PARROT using both safe and unsafe data from successful trajectories, using the hyperparameters for PARROT in Section~\ref{ap:hyperparameterdetails}.
The results given in Figure~\ref{fig:unsafedata} demonstrate that this leads to relatively higher per step safety violations, indicating that it is best to train PARROT with \textit{only} safe data from successful trajectories.

\clearpage
\pagebreak
\section{SAFER Hyperparameter Details}
\label{ap:hyperparameterdetails}
\textbf{Hyperparameter Details} We explored a number of different parameter configurations with SAFER.
We tuned $\lambda$ $(1e-4, 1e-5)$, the number of bijectors in the real NVP flow model $(3, 5)$, the number of components in the context variable $\vc$ $(8, 32, 64)$, the size of the states window $w$ $(16, 32)$, the optimizer (Adam, SGD+Momentum), and the learning rate $(1e-4, 5e-5)$.
We trained for $500$k steps and found that using a smaller number of components in the context variable led to more stable training ($8$). Setting the learning rate to $(1e-4)$ led to much quicker convergence, without sacrificing much stability. 
Furthermore, training with Adam led to divergence in some cases while SGD+Momentum tended to diverge less often. 
Between the other parameters considered, there was relatively little difference, and therefore we used a model with learning rate $1e-4$, $3$ bijectors, $8$ components, $16$ states window size, and SGD+Momentum.

\dylan{I feel like this had some more hyperparameter details in it at some point... there is also some inconsistency with SAFER wo/ constrastive loss... find + fix these TODO}
\yinlam{move the baseline section to appendix if there's no space}
\yinlam{mention which ones are for ablation studies}
\section{Baseline Methods}
\label{sec:baselines}
We select several baseline methods to compare with SAFER. We mainly focus on methods leverage action primitives trained with offline data to improve efficiency, e.g., PARROT, Prior-Explore. While we are aware of additional baseline methods, e.g., TrajRL \citep{shankar2020learning, fox2017multi}, HIRL \citep{ghadirzadeh2020data}, in the literature, we omit their comparisons here because it has been shown in prior work \citep{Singh2021ParrotDB} that their performance is consistently below that of the state of the art.  
 
\textbf{Soft Actor Critic:}  Soft-actor critic (SAC) \citep{Haarnoja2018SoftAO} is one of the standard model-free policy-gradient based RL methods. Here without using any action primitives we apply SAC to learn a policy that directly maps states in $\mathcal X$ to actions in $\mathcal A$. Later we also use SAC in all our action primitive based RL methods (e.g., SAFER, PARROT) to optimize the high-level policy. Therefore, one can view the SAC baseline as one ablation study as well.
We use the implementation from TF-Agents \citep{TFAgents}.
We used SAC with autonmatic entropy tuning and tune the number of target network update period, discount factor, policy learning rate, and Q-function learning rate.

\textbf{PARROT:} We compare against the state-of-the-art primitive learning RL method PARROT, proposed by \citet{Singh2021ParrotDB}.
Similar to SAFER, PARROT leverages a conditional normalizing flow and to train a behavioral prior using data from successful rollouts.
To enforce safety in the PARROT agent, we additionally limit the training data of its behavioral prior to \textit{both} safe and successful rollouts, otherwise PARROT may encourage unsafe behaviors.
We tune the number of bijectors in the conditional normalizing flow for PARROT ($5$, $3$), the number hidden units in each bijector layer ($128$, $256$), the learning rate ($1e-4$, $5e-5$, $1e-5$), the optimizer (Adam or SGD+Momentum), and train for $500$k steps. We find using $3$ bijectors with learning rate $1e-4$, and the Adam optimizer works best.

\textbf{Prior-Explore:} We also consider the prior-explore method proposed in \citet{Singh2021ParrotDB} as one of our baseline method.
Here the prior-explore policy combines the mapping $f_\phi$ action policy in \Eqref{eq:realnvploss} with an SAC agent to aid exploration of the RL agent. It selects an action from the prior policy with probability $\delta$ and from the SAC agent otherwise.
Followed from \citet{Singh2021ParrotDB}, we set this probability $\delta$ to $0.9$ and use mapping $f_\phi$ trained for SAFER.

\textbf{Contextual PARROT (SAFER Without Contrastive Loss):} As one ablation study we consider SAFER \textit{without} the contrastive loss. 
This setup also models the behavioral prior policy with a conditional normalizing flow and the latent safety variable but trains that only with safe and successful data. 
Note that this baseline method is equivalent to PARROT, with a policy that is a function of the latent safety variable.
We use the same parameters as PARROT with this baseline and $8$ components in safety variable because we found this number of components to be the most successful with SAFER.

\section{Training SAFER}
\label{ap:safer}
In this appendix, we provide psuedo code for the SAFER training procedure in Algorithm~\ref{alg:trainingsafer}.

  \begin{algorithm}[H]
    \caption{SAFER Training}
    \label{alg:trainingsafer}
    \begin{algorithmic}
      \REQUIRE SAFER Behavioral Prior $f_\phi$, Safety Variable Posterior $q_\rho ( \vc | \Lambda )$, safe dataset $\mathcal{D}_{\textrm{safe}}$, unsafe dataset $\mathcal{D}_{\textrm{unsafe}}$, Steps $N$, $\lambda$
     \STATE Let \texttt{flow\_loss($\cdot$)} refer to Equation~\ref{eq:realnvploss}
      \FOR {$n = 1,...,N$}
        \STATE $(\vs, \va, \Lambda )_{\textrm{Safe}} \sim \mathcal{D}_{\textrm{Safe}}$ \hfill \algorithmiccomment{{\color{magenta}\texttt{Sample safe + unsafe batches of data}}     }
        \STATE $(\vs, \va, \Lambda )_{\textrm{Unsafe}} \sim \mathcal{D}_{\textrm{Unsafe}}$ 
        \STATE $\vc_{\textrm{Safe}} \sim q_\rho(c | \Lambda_{\textrm{Safe}})$ \hfill\algorithmiccomment{{\color{magenta}\texttt{Sample safety variables}}     }
        \STATE $\vc_{\textrm{Unsafe}} \sim q_\rho(c | \Lambda_{\textrm{Safe}})$
        \STATE $\mathcal{L}_{\textrm{safe}} \leftarrow $ log (\texttt{flow\_loss}$( \vs_{\textrm{safe}}; \va_{\textrm{safe}}; \vc_{\textrm{safe}} )) $ \hfill\algorithmiccomment{{\color{magenta}\texttt{Compute log-likelihoods}}}
        \STATE $\mathcal{L}_{\textrm{unsafe}} \leftarrow $ log (\texttt{flow\_loss}$( \vs_{\textrm{unsafe}}; \va_{\textrm{unsafe}}; \vc_{\textrm{unsafe}} )) $
        \STATE $D_{\textrm{KL}}^{\textrm{Safe}} \leftarrow  D_{\textrm{KL}} \left( q_\rho(\vc | \Lambda_{\textrm{Safe}}) || p(\vc) \right)$  \hfill\algorithmiccomment{{\color{magenta}\texttt{Compute KL of safety variables}}}
        \STATE $D_{\textrm{KL}}^{\textrm{Unsafe}} \leftarrow  D_{\textrm{KL}} \left( q_\rho(\vc | \Lambda_{\textrm{Unsafe}}) || p(\vc) \right)$ 
        \STATE $NLL \leftarrow - ( \mathcal{L}_{\textrm{safe}} - \lambda \cdot  \mathcal{L}_{\textrm{unsafe}} - D_{\textrm{KL}}^{\textrm{Safe}} - D_{\textrm{KL}}^{\textrm{Unsafe}} )$
        \STATE Minimize $NLL$ and update $\phi, \rho$ \hfill\algorithmiccomment{{\color{magenta}\texttt{Update SAFER}}}
      \ENDFOR
        \STATE \textbf{Return:} SAFER Behaviors Prior $f_\phi$, Safety Variable Posterior $q_\rho (\vc | \Lambda)$
    \end{algorithmic}
  \end{algorithm}

\section{Setting the Safety Assurance}
\label{ap:safetyassurances}
In this appendix, we provide psuedo code for the SAFER safety assurances procedure in Algorithm~\ref{alg:safeassure}.
This algorithm provided a numerical gradient-free approach to find an optimal bound $\eta$ that included $\epsilon$ portion safe actions.

\newlength\myindent
\setlength\myindent{2em}
\newcommand{\bindent}{%
  \begingroup
  \setlength{\itemindent}{\myindent}
  \addtolength{\algorithmicindent}{\myindent}
}
\newcommand{\eindent}{\endgroup}



      \begin{algorithm}[H]
        \caption{SAFER Safety Assurances}
        \label{alg:safeassure}
        \begin{algorithmic}
          \REQUIRE Initial bound $\eta_0$, Desired percent safe actions $\epsilon$, SAFER prior $f_\phi$, Safe dataset $\mathcal{D}_{\textrm{safe}}$, Unsafe dataset $\mathcal{D}_{\textrm{unsafe}}$
          \STATE \textbf{define} 
          \FUNCTION{get\_in\_bound(dataset $\mathcal{D}$, bound $\eta_t$)} 
          \STATE \texttt{// This function computes the abstract actions $\vz$ within bound $\eta_t$}
          \STATE $\mathcal{Z}$ $\leftarrow$ \{\}
          \FOR{($\vs$, $\va$, $\Lambda$) in $\mathcal{D}$}
          \STATE \texttt{// Iterate over tuple (state $\vs$, action $\va$, and context $\Lambda$)}
            \STATE $\vc \leftarrow \mathbb{E}_{q_\rho (\cdot | \Lambda)} \left[ \vc \right]$, $\vz$ $\leftarrow$ $f_{\phi}^{-1} (\va ; \vs; \vc) $
             \hfill \algorithmiccomment{{\color{magenta}\texttt{Get abstract action $\vz$ from $\vs$, $\va$, and $\Lambda$}} }
            \IF{$\vz$ within bound $\eta_t$}
                \STATE $\mathcal{Z}$ = $\mathcal{Z}$ $\cup \; \vz$
                \hfill \algorithmiccomment{{\color{magenta}\texttt{Add $\vz$ if its within bound $\eta_t$}} }
            \ENDIF
            \ENDFOR 
            \STATE \textbf{return} $\mathcal{Z}$
         \ENDFUNCTION
         \STATE $\eta$ $\leftarrow$ $(- \eta_0, \eta_0)$ 
         \hfill \algorithmiccomment{{\color{magenta}\texttt{Initialize bound}} }
         \STATE \texttt{done} $\leftarrow$ False
         \WHILE{not \texttt{done}}
            \STATE $\mathcal{Z}_{\textrm{safe}}^{\eta}$ = get\_in\_bound($\mathcal{D}_{\textrm{safe}}$, $\eta$), $\mathcal{Z}_{\textrm{unsafe}}^{\eta}$ = get\_in\_bound($\mathcal{D}_{\textrm{unsafe}}$, $\eta$)
            \hfill \algorithmiccomment{{\color{magenta}\texttt{Get $\vz$ in current bound $\eta$}} }
            \STATE $S$ $\leftarrow$ $|\mathcal{Z}_{\textrm{safe}}^{\eta}|$ + $|\mathcal{Z}_{\textrm{unsafe}}^{\eta}|$
            \STATE $\mathcal{Z}_{\textrm{safe}}^{\epsilon} \sim$ sample $\floor{S \times \epsilon}$ items from $\mathcal{Z}_{\textrm{safe}}^{\eta}$,  $\mathcal{Z}_{\textrm{unsafe}}^{1-\epsilon} \sim$ sample $\floor{S \times (1-\epsilon)}$ items from $\mathcal{Z}_{\textrm{unsafe}}^{\eta}$ 
            \STATE $\eta \; \leftarrow $ the max component absolute value across $\mathcal{Z}_{\textrm{unsafe}}^{1-\epsilon}$ and $\mathcal{Z}_{\textrm{safe}}^{\epsilon}$
            \hfill \algorithmiccomment{{\color{magenta}\texttt{Update bound}} }
            \IF{$\epsilon$ portion of items across $\mathcal{D}_{\textrm{safe}}$ and $\mathcal{D}_{\textrm{unsafe}}$ within bound ($-\eta$, $\eta$) are safe}
                \STATE \texttt{done} $\leftarrow$ True
                \hfill \algorithmiccomment{{\color{magenta}\texttt{Break if bound $\eta$ contains desired portion safe actions}} }
            \ENDIF
         \ENDWHILE
         \STATE \textbf{return} $\eta$
        \end{algorithmic}
      \end{algorithm}

\section{Additional Task Examples}

\begin{figure}
    \centering
    \includegraphics[width=\textwidth]{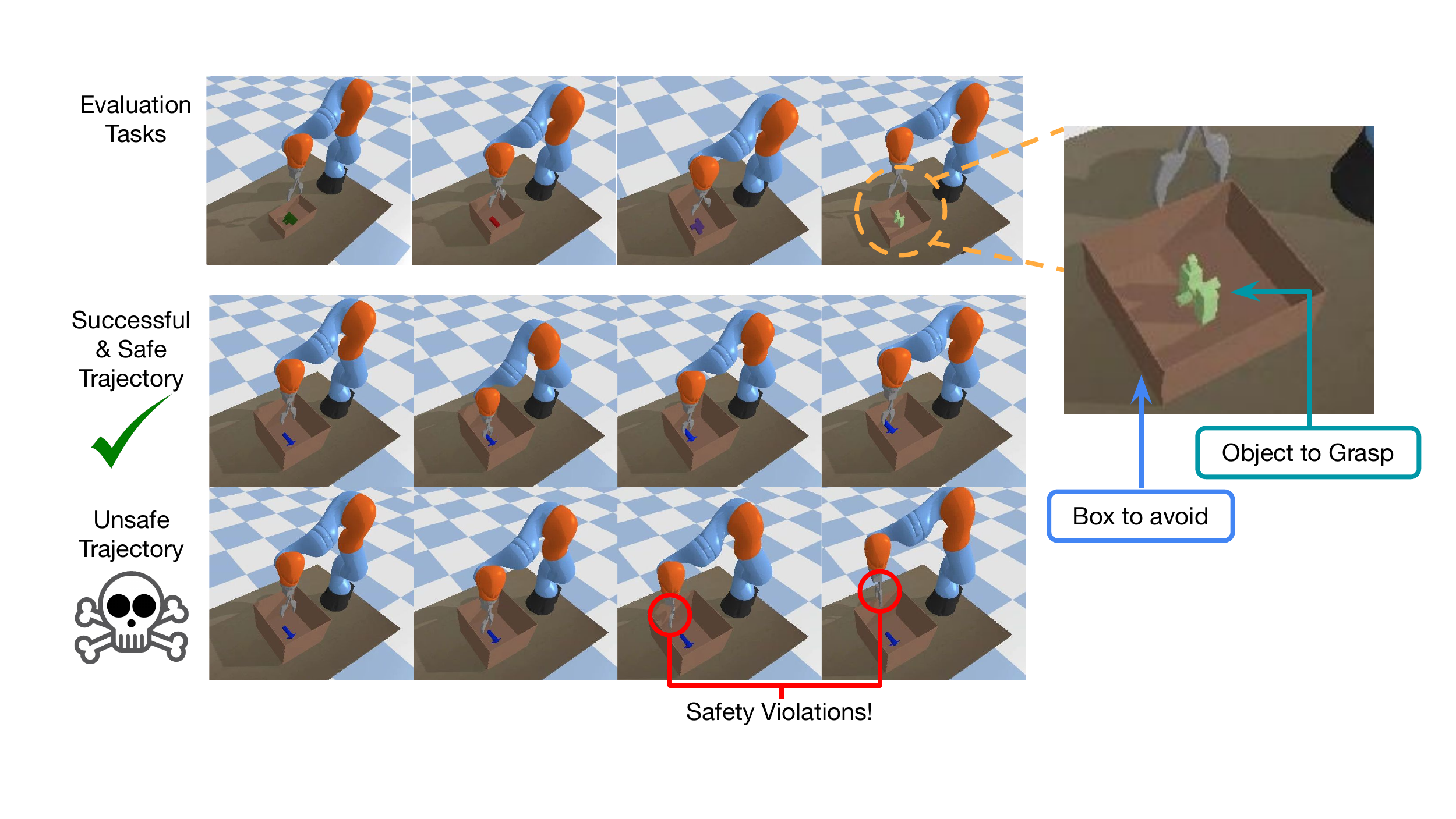}
    \caption{Additional examples of tasks included in the safe robotic grasping environment (top row). The tasks all use different sizes containers, to represent different difficulties in preserving safe behavior. We also provide a zoomed in version of the task (right hand side). Finally, we also include the examples of safe and unsafe trajectories provided in the main paper (Figure~\ref{fig:tasks}) for completeness}
    \label{fig:add_task_examples}
\end{figure}

In this Appendix, we provide additional examples of the tasks included in the safe robotic grasping environment in Figure~\ref{fig:add_task_examples}.

\end{document}